\documentclass{article}

    \usepackage[final, nonatbib] {neurips_2023}

\usepackage[utf8]{inputenc} % allow utf-8 input
\usepackage[T1]{fontenc}    % use 8-bit T1 fonts
\usepackage{hyperref}       % hyperlinks
\usepackage{url}            % simple URL typesetting
\usepackage{booktabs}       % professional-quality tables
\usepackage{amsfonts}       % blackboard math symbols
\usepackage{nicefrac}       % compact symbols for 1/2, etc.
\usepackage{microtype}      % microtypography
\usepackage{xcolor}         % colors
\usepackage{graphicx}       % add graphcs
\usepackage{amsmath}        % math
\usepackage{amssymb}

\title{Exploiting Contextual Structure to Generate Useful Auxiliary Tasks}

\author{%
  Benedict Quartey\\
  Department of Computer Science\\
  Brown University\\
  \And
  Ankit Shah \\
  Department of Computer Science\\
  Brown University\\
  \AND
  George Konidaris \\
  Department of Computer Science\\
  Brown University\\
}

\begin{document}
\maketitle
\begin{abstract}
Reinforcement learning requires interaction with environments, which can be prohibitively expensive, especially in robotics. This constraint necessitates approaches that work with limited environmental interaction by maximizing the reuse of previous experiences. We propose an approach that maximizes experience reuse while learning to solve a given task by generating and simultaneously learning useful auxiliary tasks. To generate these tasks, we construct an abstract temporal logic representation of the given task and leverage large language models to generate context-aware object embeddings that facilitate object replacements.  Counterfactual reasoning and off-policy methods allow us to simultaneously learn these auxiliary tasks while solving the given target task. We combine these insights into a novel framework for multitask reinforcement learning and experimentally show that our generated auxiliary tasks share similar underlying exploration requirements as the given task, thereby maximizing the utility of directed exploration. Our approach allows agents to automatically learn additional useful policies without extra environment interaction. 
\end{abstract}

\section{Introduction}
Reinforcement learning (RL) is a general-purpose paradigm that models agents interacting with environments. It has proven to be a robust approach to sequential decision-making and has recently seen several exciting successes \cite{mnih2015human,schulman2017proximal,berner2019dota}. However, to learn valuable behaviors that accomplish tasks, an agent must explore by repeatedly interacting with its environment, which can be prohibitively expensive, especially in robotics. Therefore it is imperative to make efficient use of experience data. One approach is to exploit the fact that, while exploring to solve any given task, an agent acquires environmental experiences that could be valuable for learning to solve many closely related tasks. Consider the task of learning to make tea in an unfamiliar house. To solve this task, a simple sequence of goals could be: \emph{get to the kitchen,} \emph{get a cup}, \emph{get a teabag}, \emph{get water from the faucet}, and \emph{get milk from the fridge}. Once an agent discovers the location of a fridge while learning to solve this task, it should be able to reuse that experience to solve a related task, such as \emph{get a drink from the fridge}.

\begin{figure}
\centering
  \includegraphics[scale=0.29]{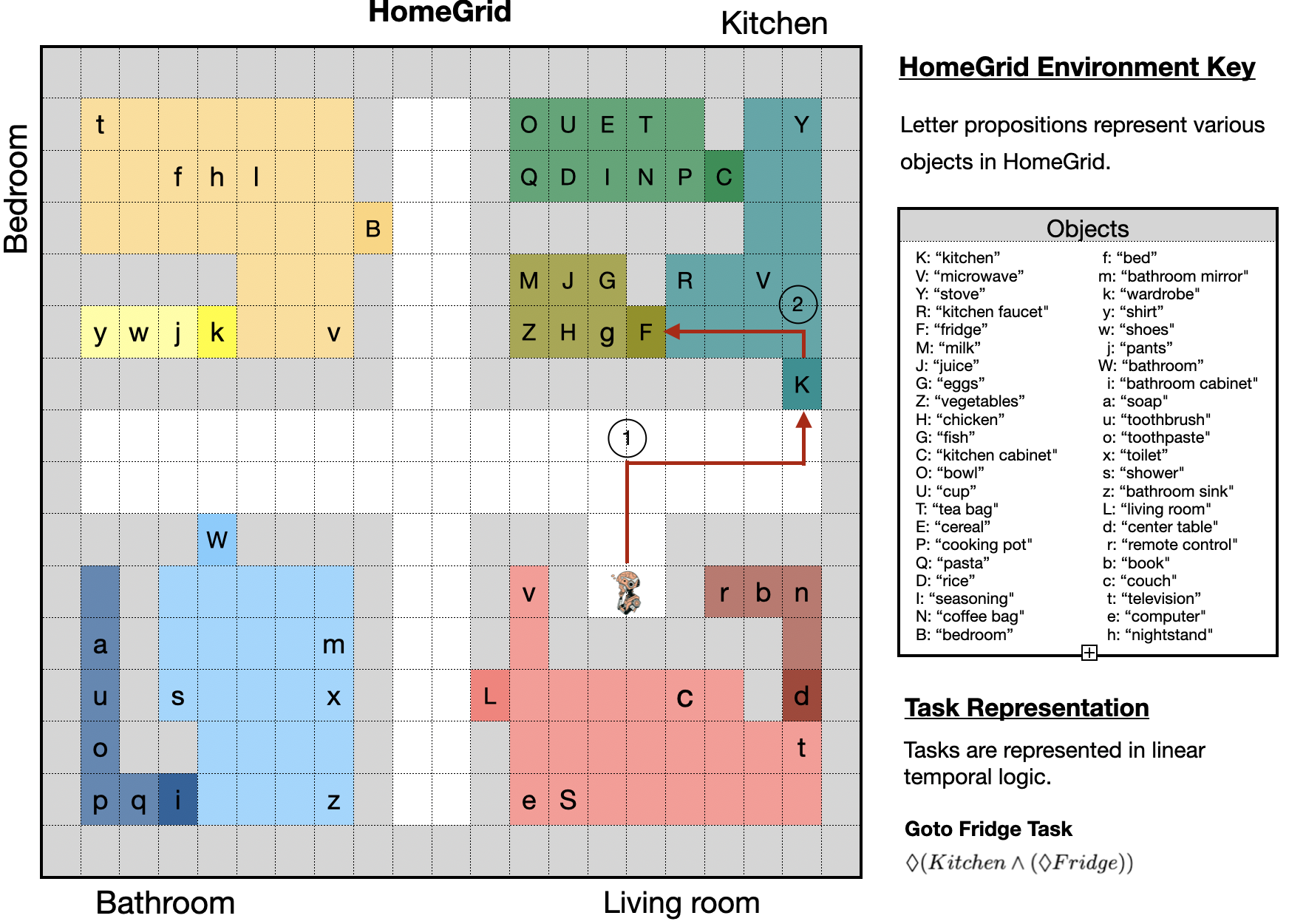}
  % \vspace*{-4mm}
  \caption{HomeGrid, a deterministic discrete grid-world domain. Agents in this world complete tasks by visiting grid locations corresponding to objects in the environment. Tasks are specified with LTL formulae that represent the sequence/ordering of subgoals necessary for completing a given task. Satisfying these tasks involves visiting relevant grid cells in an acceptable order determined by the task specification. As an example, the numbered arrows in the diagram indicate a policy for satisfying the \textbf{Goto Fridge} task. 
  % A complete list of letter propositions and their descriptions can be found in the Appendix section.
  }
  \label{fig:environment}
  \vspace*{-3mm}
\end{figure} 

Prior experience replay works \cite{lin1992self,schaul2015prioritized,andrychowicz2017hindsight} introduce effective methods of reusing previous experiences, and off-policy learning algorithms \cite{icarte2022reward,icarte2018using,rodrigoLPOPL} also enable cross-task learning; the agent can use data generated by a \emph{behavior policy} to learn to perform a different \emph{target policy}. However, these approaches are limited in their ability to specify auxiliary tasks that maximally benefit from counterfactual experience reasoning and off-policy learning, in part because they do not make any assumptions about task structure.

We posit that task structure, as well as semantic and contextual structure in object-centric environments, can be exploited to generate auxiliary tasks. We propose a new method TaskExplore, that given a target task, uses temporal logic expressions as a means for generating contextually similar auxiliary tasks, by swapping objects using context-aware embeddings generated by large language models. In this setting, we can leverage counterfactual reasoning and off-policy methods to simultaneously learn these auxiliary tasks while learning the given task, with a behavior policy only conditioned on the given task. Our approach maximizes the utility of directed exploration experience, particularly in complex environments that require agents to constrain/direct their exploration. We show empirically that auxiliary tasks generated by TaskExplore maximally leverage the directed experience of a single-task curriculum.

In summary, we present two main contributions in this paper:
\begin{enumerate}
    \item We present a method of using context-aware object embeddings and abstract temporal logic task representations to generate useful auxiliary tasks that share underlying exploration requirements with a given target task.
    \item We demonstrate empirically that this class of generated tasks results in better experience transfer than randomly generated tasksets and uniquely benefits from directed exploration on the primary task. 
\end{enumerate}

\section{Background}
In this work, we focus on object-centric environments 
%\cite{diuk2008object} 
and leverage linear temporal logic \cite{pnueliLTL} to describe temporally extended tasks involving objects. These tasks are learned using off-policy learning for linear temporal logic \cite{rodrigoLPOPL}.
% We leverage the structure inherent to linear temporal logic to represent temporal tasks and object attributes
% To generate auxiliary tasks, we leverage structure in terms of task specifications and attributes of objects present in environment

% \subsection{Object Oriented Markov Decision Process}
% An Object-oriented Markov decision process (OOMDP) \cite{diuk2008object} is a Markov decision process where its state space is structured into objects, and its transition function focuses on interactions between objects. A given environment in the OOMDP paradigm augments the state, action, transition function, reward function and discount factor tuple of regular MDPs with a set of objects $O = \{o_1,...,o_n\}$, where each object is an instance of a class from a predefined set of classes $C = \{C_1,..., C_m\}$. Each class typically also includes a set of attributes $Att(C) = \{C.a_1,..., C.a_k\}$.

\subsection{Linear Temporal Logic}
Linear Temporal Logic (LTL) \cite{pnueliLTL}, presents an expressive grammar to specify temporal behavior. LTL expressions/formulae are composed of atomic propositions, the logical connectives: negation ($\neg$), conjunction ($\land$), disjunction ($\lor$), implication ($\rightarrow$); and a set of temporal operators: next ($\bigcirc$), until (\text{U}), always ($\square$) and eventually($\lozenge$). The minimal syntax of an LTL formula is defined below:
\begin{equation}
    \varphi := p\ |\ \neg\varphi_1\ |\ \varphi_1\wedge\varphi_2\ |\ \bigcirc\varphi\ |\ \varphi_1 \text{U} \varphi_2
\end{equation}
where $p$ is an atomic proposition, a boolean literal that captures a property of the environment; $\varphi_1$ and $\varphi_2$ are valid LTL formulas. LTL formulas present an alternative, often more expressive and natural way of specifying reward objectives in the reinforcement learning setting \cite{littman2017environment}. They allow the expression of explicit specifications that characterize the successful execution of a task. Consider a sequential navigation task of visiting a kitchen and then visiting a fridge, where \emph{Kitchen} and \emph{Fridge} are Boolean atomic propositions that can be observed. The LTL formula below can be used to specify this task:
\begin{equation}
    \lozenge(Kitchen \land (\lozenge Fridge))
\end{equation}

LTL formulas can be progressed given a sequence of truth assignments of propositions \cite{bacchus2000using} to determine which parts of the formula have been satisfied by the states seen so far and which parts remain. This is particularly useful in the reinforcement learning domain as this provides a method of tracking non-Markovian objectives. In fact, a class of reinforcement learning algorithms such as geometric-LTL (G-LTL) \cite{littman2017environment}, Q-learning for Reward machines (Q-RM) \cite{icarte2018using} and LPOPL \cite{rodrigoLPOPL} leverage this to construct and solve a product MDP using the environment state space and automaton representation of LTL specifications. 
% Useful objectives specified in ltl .... also works exploring natural language to LTL [cite] such that it is easier to use ltl as a reward function from just giving natural instructions.

\subsection{Off-policy Learning with LTL}
Off-policy RL methods address the setting where an agent learns a desired \emph{Target Policy} while interacting in the environment using a different \emph{Behavior Policy}. LTL Progression for Off-Policy Learning (LPOPL) \cite{rodrigoLPOPL} adapts the Q-learning off-policy algorithm \cite{watkins1992q} to simultaneously learn policies for multiple LTL tasks during the same environment interaction. 

Given a set of LTL task specifications $\phi$, LPOPL first extracts subtasks from each specification via LTL progression~\cite{bacchus2000using}, ie. progressing every given formula over all possible truth assignments of the set of environment propositions. LPOPL then iteratively performs a series of episodes and learns a separate Q-value function for each task and extracted subtasks (also expressed as LTL formulae). At each iteration, a task is selected from $\phi$ and used as the objective of the episode and actions are selected with an epsilon greedy behavior policy conditioned on that task. However, all Q-value functions are updated at each step via off-policy updates, allowing the agent to make progress on tasks that may not be its current objective. 

To help describe LPOPL, consider an example with the given set of tasks $\phi = \{ \lozenge(a \land \lozenge b),  \lozenge(c \land \lozenge b)\}$. Progressing both tasks would result in extracting the subtask $\lozenge b$. LPOPL will then initialize three Q-value functions $Q_{\lozenge(a \land \lozenge b)}$, $ Q_{\lozenge(c \land \lozenge b)}$ and $Q_{\lozenge b}$. To run an episode, a task will be selected from $\phi$, say $ \lozenge(a \land \lozenge b)$, then actions will be selected epsilon greedy on $Q_{\lozenge(a \land \lozenge b)}$. Once the proposition \emph{a} becomes \emph{true} during the episode, the behavior policy becomes epsilon greedy on $Q_{\lozenge b}$, since progressing  $\lozenge(a \land \lozenge b)$ with an assignment of True for the proposition \emph{a} transforms the LTL formula to $\lozenge b$. This leads to a desirable property of LPOPL: task specifications that share progressed LTL forms can share subtask policies.

% remove a ton and end with this distinction
We borrow LPOPL's subtask extraction and Q-value function update strategies to accelerate off-policy learning of multiple tasks. However, our  approach differs  as we do not learn from a curriculum of tasks. We instead use a behavior policy conditioned on a \emph{single} given task and apply counterfactual reasoning on experiences from this task to simultaneously solve generated auxiliary tasks.

\section{Related Work}
Experience replay methods consider single-task curriculum problems where the cardinality of the set of tasks used in extracting samples for transfer is one (1) and includes only the target task \cite{narvekar2020curriculum}. The focus in these works is discovering optimal methods of organizing and training on the experience acquired from single tasks. Prioritized experience replay \cite{schaul2015prioritized} improves on Experience Replay \cite{lin1992self}, which uniformly sampled from a replay memory, by prioritizing important transitions so they are sampled more frequently. Hindsight experience replay (HER) \cite {andrychowicz2017hindsight} employed exploration as an implicit curriculum and introduced learning from alternate realizations from experiences in the replay memory, by relabelling experiences based on goals that were actually achieved rather than what the agent was aiming to achieve. HER's counterfactual experience reuse is limited to singular goal states, and so cannot encode expressive temporally extended behaviors such as reaching a goal state while encountering specific intermediary states \cite{icarte2022reward}. Additionally, HER's alternative goals are intermediate  samples of the target task and not distinct alternative tasks. 

Other works such as Counterfactual experiences for reward machines (CRM) \cite{icarte2022reward}, Q-learning for Reward Machines \cite{icarte2018using} and LPOPL \cite{rodrigoLPOPL} introduce algorithms for applying off-policy updates to simultaneously learn multiple action-value functions. They address HER's limitation in applying counterfactual experience to temporally extended alternative tasks. However, in these works, the alternative tasks that benefit from counterfactual reasoning are assumed to be known or given. Additionally, behavior policies that dictate environment interactions from which counterfactual experiences are generated are conditioned on curricula consisting of multiple tasks–typically including the alternative tasks that benefit from these synthetic experiences. 

% A gap in these approaches is in problems where an agent has a single-task curriculum \cite{narvekar2020curriculum} which is its target task and may not have the luxury of additional environment interaction following a curriculum of multiple tasks to generate diverse experience samples, from which counterfactual experiences may be generated. In environments with large and complex state spaces, such agents require some form of directed exploration strategy toward the given task to acquire relevant experiences and overcome sample inefficiency.

Our work seeks to present a solution to how agents might automatically generate expressive temporally extended auxiliary tasks–--in contrast to intermediate tasks–--that can maximally leverage the directed experience of a single-task curriculum in object-centric environments.
% Structure in object-oriented environments can be exploited towards this end .. Temporal logics on the other hand present
TaskExplore is distinct from the limited task generation works in curriculum learning literature \cite{narvekar2016source, silva2018object}, where tasks are not manually designed. The goal of our approach is not to generate good intermediate tasks to obtain experience samples, as is the goal in task generation for curriculum learning \cite{narvekar2020curriculum}. Our focus is rather to generate distinct auxiliary tasks that maximally leverage directed experience from single-task curricula, which is more akin to life-long learning contexts, where agents learn to generalize from very small or constrained datasets \cite{thrun1998lifelong}.

\begin{figure*}
\centering
 \makebox[\textwidth][c]{\includegraphics[width=\textwidth]{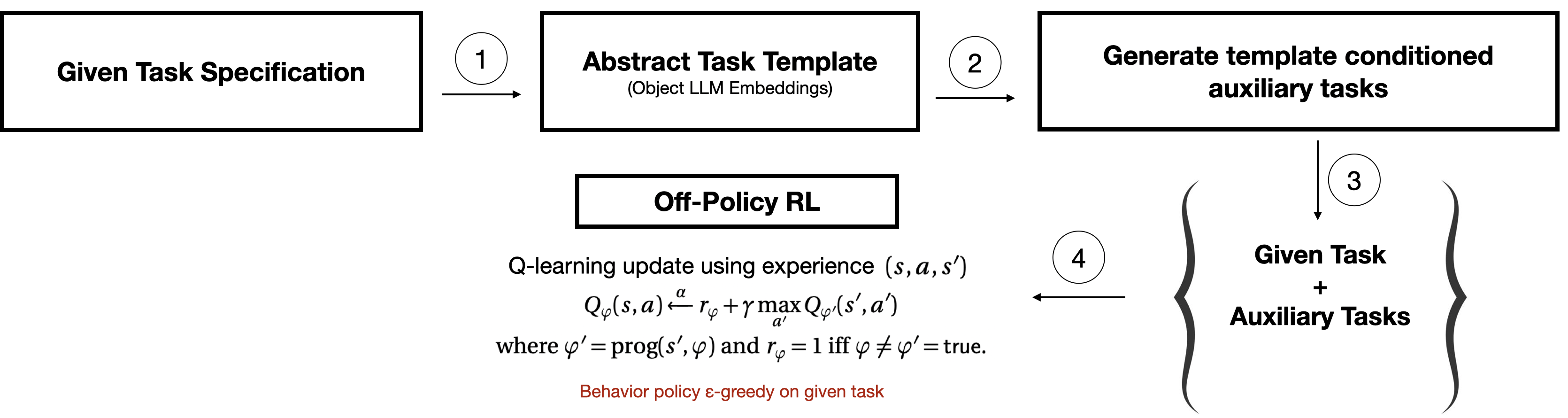}}%
   % \vspace*{-5mm}
  \caption{This figure depicts the TaskExplore framework. Given a task specified in linear temporal logic, we construct an abstract task template that replaces instance object propositions in the given formula with large language model embeddings of their descriptions, capturing various relevant attributes of each object. We then generate auxiliary tasks by selecting objects from the environment for each proposition node in our abstract task template using the cosine similarity metric. We initialize policies (Q-value functions) for the given task and all auxiliary tasks and perform RL where actions are selected $\epsilon$-greedy on only the given task, gathering directed experiences necessary for solving the given task. At each learning step, all Q-value functions are updated via off-policy Q-learning updates.}
  \label{fig:approach}
    \vspace*{-5mm}
\end{figure*}

\section{Problem Definition}
To define our problem formally we propose Object Oriented Non-Markovian reward decision process (OO-NMRDP), combining ideas from the Object-Oriented Markov decision process (OOMDP) \cite{diuk2008object} and Non-Markovian reward decision process (NMRDP) \cite{brafman2018ltlf, camacho2017non, thiebaux2006decision} formalisms. An OO-NMRDP is an 8-tuple $M =⟨O, C, L, S, A, T, R_\varphi, \gamma⟩$, where O is a set of Boolean propositions representing objects present in the environment and detectable by a labeling function $L : S \rightarrow 2^O$ that maps states to these Boolean propositions, specifying which propositions are true in which states. C is the set of object classes, S is the set of states, A is a set of actions, $\gamma \in [0, 1]$ is the discount factor and $T: S \times A \times S \rightarrow [0, 1]$ represents the transition dynamics of the environment. Unlike regular MDPs the reward function $R_\varphi$ is defined over state histories, where the agent receives a reward of 1 if and only if the sequence of seen states in a given episode satisfies the LTL formula $\varphi$.
% a binary signal indicating successful task completion
% ie.
  \begin{equation}
        R_\varphi (\langle s_0,...,s_n\rangle) = 
        \begin{cases}
            1 & ~ \mathrm{if}~ \delta_{0:n-1} \nvDash \varphi ~\text{and}~ \delta_{0:n} \models \varphi \\
            0 &~  \mathrm{otherwise},
        \end{cases}
    \end{equation}
where $\delta_{i:j} = \langle L(s_i),...,L(s_j)\rangle$. The learning agent does not have access to either the set of object classes $C$ or the transition dynamics of the environment $T$. We express sequential~\cite{littman2017environment} or soft ordering constraint tasks~\cite{liu2022skill} as LTL formulas over the set of propositions O. 

% The learning agent does not have access to the true set of object classes, as such we construct and cluster semantic context-aware representations of objects, which represent emergent object classes.

\begin{figure*}
\centering
 \makebox[\textwidth][c]{\includegraphics[width=\textwidth]{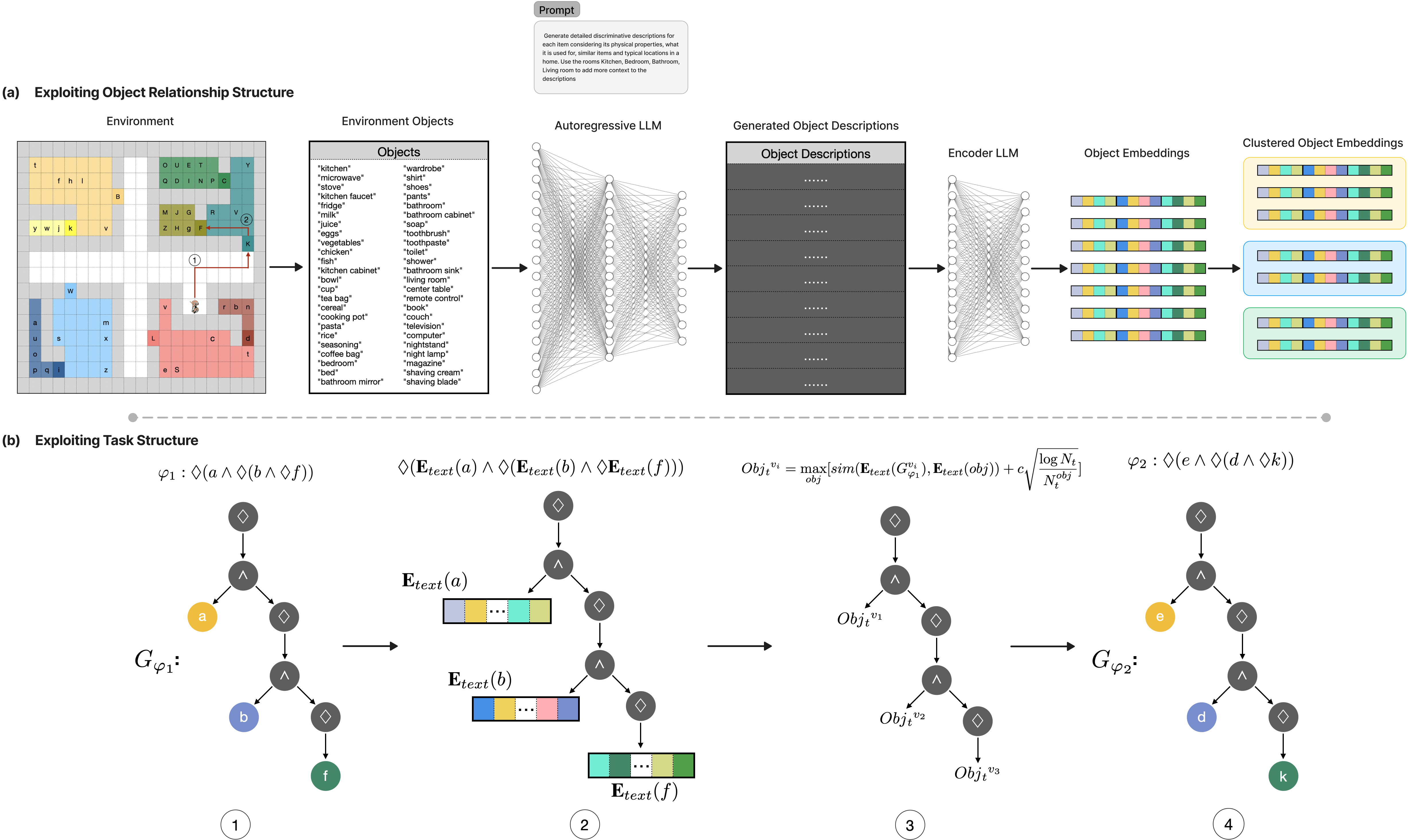}}%
  \caption{This figure depicts how TaskExplore constructs and leverages context-aware object embeddings and abstract task representations/templates. In \textbf{Figure a}, we use an autoregressive LLM to generate detailed descriptions for the list of objects in our environment and use an encoder language model to encode these generated descriptions into a 768-dimensional vector for each object. We then cluster these description embeddings, discovering object classes that capture the semantic and contextual similarity between objects. In \textbf{Figure b}, our approach constructs a task template by representing proposition nodes in the abstract syntax graph of a given LTL formula with embeddings of corresponding objects. With this task template, we can create new contextually similar tasks by selecting objects from the environment based on their cosine similarity, balancing selections between highly correlated objects and relevant yet unseen objects.
  }
     % \vspace*{-1mm}
  % Talk about getting embeddings that capture relevant attributes of objects, talk about clustering, talk about task generation}
  \label{fig:approach2}
\end{figure*}
\vspace{20mm}
\section{Exploiting contextual structure to generate auxiliary tasks}
Humans can exploit the structure inherent in object-centric environments, in terms of objects and their relationships with each other, and the compositional structure of tasks. With this structure, they are able to learn tasks while thinking of alternate ways in which the experience gathered may be useful for other tasks. This ability allows humans to gain multiple useful skills when learning any one specific thing. We present how abstract temporal logic representations of tasks and context-aware embeddings of objects in an environment could be used to equip RL agents with this ability.

Figure \ref{fig:approach} illustrates the high-level steps in our method. Given a sequential task specified in linear temporal logic $\varphi_1$, our framework leverages the compositional syntax of LTL and real-world contextual relationships between objects to develop a set of auxiliary tasks $Aux_{\varphi1}$= $\{a_1,...,a_n\}$ that possess similar underlying exploration requirements as $\varphi_1$, via object swaps. We then initialize a policy bank with Q-value functions for $\varphi_1$ and each task in the generated auxiliary task set $Aux_{\varphi1}$. We simultaneously learn a policy for $\varphi_1$, and all the auxiliary tasks using off-policy updates akin to LPOPL \cite{rodrigoLPOPL}. The agent always follows an $\epsilon$-greedy policy with respect to $\varphi_1$. A key distinction between our learning approach and LPOPL is the absence of a multi-task curriculum of target tasks and our behavior policy which is conditioned on just the given task, constraining exploration in the environment.

In section \ref{sec:objectstructure} we present a detailed look into how we construct context-aware object embeddings from our environment, these embeddings are used to determine relevant objects for swaps. Section \ref{sec:taskstructure} looks at how we use these object embeddings to construct abstract task templates from which we generate auxiliary tasks. Figure \ref{fig:approach2} visualizes these two processes. Finally, in Section \ref{sec:learning} we explain how counterfactual reasoning and off-policy learning are used to simultaneously learn policies for the generated auxiliary tasks.

\subsection{Exploiting Structure in Object Relationships}\label{sec:objectstructure}
Large language models are trained on large text corpora and encode useful common-sense and context-aware human knowledge, as such act as good priors for structuring relationships between objects in object-centric environments. Our method leverages this class of models to generate discriminative context-aware embeddings of objects. Autoregressive language models such as GPT \cite{radford2018improving} are adept at language generation as they are trained to maximize the likelihood of the next token given previous tokens. Encoder-decoder and encoder-only large language models such as T5 \cite{raffel2020exploring} and BERT \cite{devlin2018bert} on the other hand are adept at generating compact representations that effectively capture sentence context and semantics. We use the InstructGPT \emph{text-davinci-003} model \cite{ouyang2022training} to generate detailed descriptions for the list of objects in our environment and use the \emph{Sentence-T5} encoder model \cite{ni2021sentence} to encode these generated descriptions into a 768-dimensional vector for each object. We then cluster these description embeddings using the K-means algorithm \cite{lloyd1982least,macqueen1965some} with K-means++ initialization \cite{arthur2006k}.

\begin{figure*}
\centering
 \makebox[\textwidth][c]{\includegraphics[width=\textwidth]{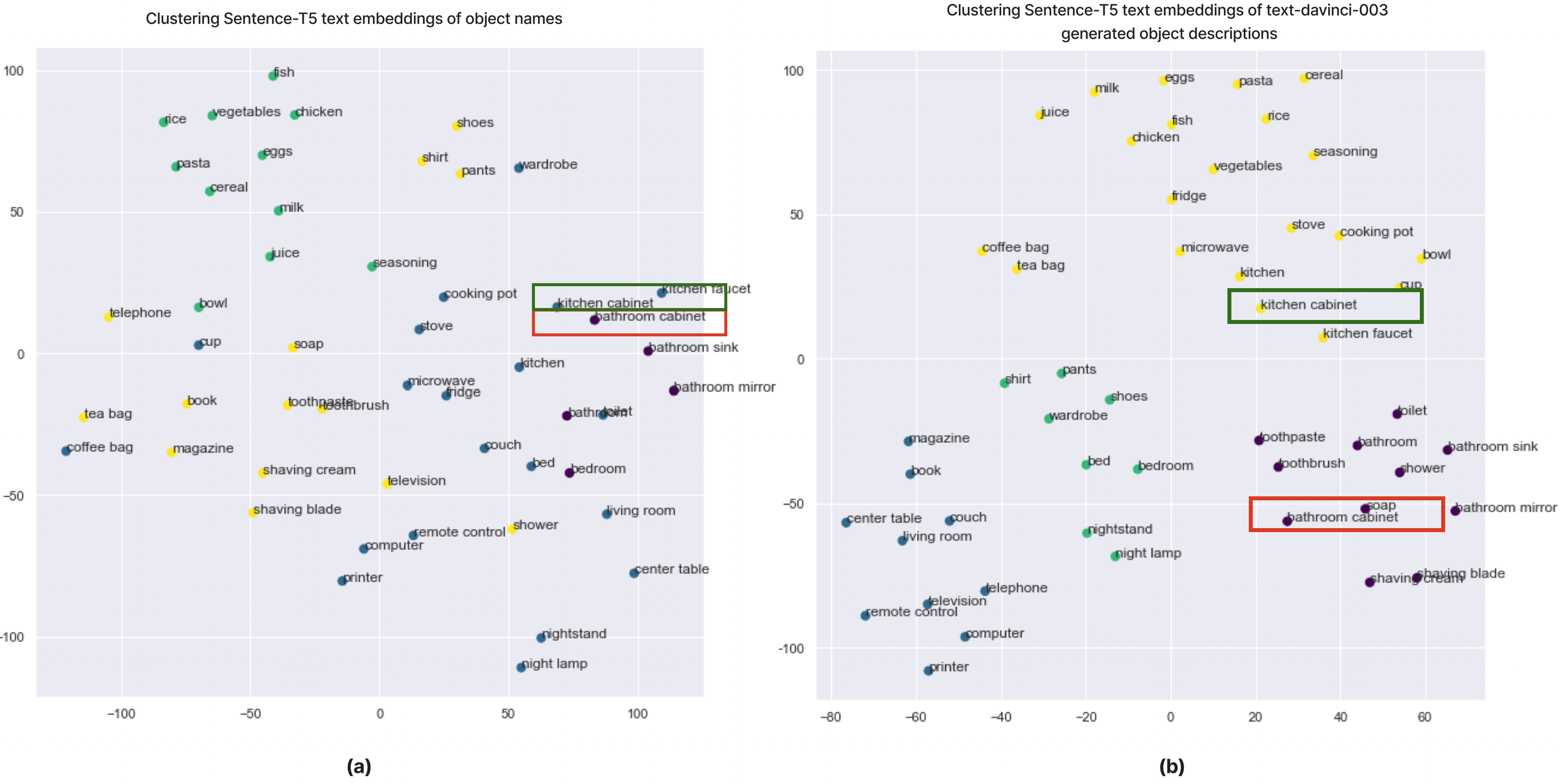}}%
      % \vspace*{-5mm}
  \caption{This figure depicts the results of performing k-means clustering on 768-dimensional embedding vectors for each environment object, results are visualized in a 2D latent space. Embeddings for each object in Figure (a) are generated by encoding the shown object name using the Sentence-T5 model. Conversely, embeddings in Figure (b) are generated by Sentence-T5 encoding text descriptions of each object generated by text-davinci-003. The number of clusters used in the k-means algorithm was four(4) based on the number distinct exploration zones in HomeGrid. \textbf{Note that embeddings generated from LLM object descriptions improved the separation of emergent cluster boundaries, and desirably increased the distance in latent space between similar yet contextually different objects such as Kitchen Cabinet and Bathroom Cabinet.}}
  \label{fig:clustering}
       \vspace*{-2.5mm}
\end{figure*}

As an ablation, we investigate Sentence-T5's ability to generate context-aware embeddings for clustering based on object name alone, shown in Figure \ref{fig:clustering}a. Figure \ref{fig:clustering}b alternatively shows the improvement in clustering results using descriptions generated by the \emph{text-davinci-003} model, presenting insights into the benefits of using large language models for context-aware data augmentation for downstream tasks.

\subsection{Exploiting Structure in Task Composition}\label{sec:taskstructure}
We leverage the inherent compositionality of LTL to create generalizable representations or templates of tasks given an instance LTL formula. This task template allows us to generate related auxiliary tasks. Similar to the representation format used in prior works \cite{vaezipoor2021ltl2action} we parse a given formula $\varphi_1$ into its abstract syntax tree and represent it as a directed graph $G_{\varphi_1} = (V_{\varphi_1}, E_{\varphi_1})$. Edges $E_{\varphi_1}$ in this graph connect parent operators to their subformulas and Vertices $V_{\varphi_1}$ represent nodes that are either operators or atomic propositions, as shown in Figure \ref{fig:approach2}b.1. This representation allows us to efficiently exploit the structure of the given task in constructing auxiliary tasks.

We create an abstract task template by traversing $G_{\varphi_1}$ and replacing instance proposition nodes--–which represent objects in our environment--–with embeddings of their descriptions as shown in Figure \ref{fig:approach2}b.2. With this abstract task template, we generate $x$ new auxiliary tasks by swapping relevant objects from the environment at each proposition node. We select objects whose embedding lies in the same embedding cluster and prioritize those with the highest cosine similarity with the template embedding node being considered. We introduce a simple value-dependent object selection metric based on upper confidence bounds \cite{auer2002using} that balances out selecting high cosine similarity objects and relevant but unseen objects.

% that balances out selecting unseen relevant objects and max cosine similarity high-value objects and relevant but unseen objects.

Equation \ref{eq:obj_ucb} governs object selection for each embedding node in a given template.
\begin{equation} \label{eq:obj_ucb}
    Obj{_t}^{v_i} = \max_{obj}[ sim(\textbf{E}_{text}(G_{\varphi{_1}}^{v_i}),\textbf{E}_{text}(obj)) + c \sqrt{\frac{\log{N_t}}{N_{t} ^{obj} }}]
\end{equation}
where \textbf{$Obj{_t}^{v_i}$} is the chosen object proposition for template node $i$ at trial t; $sim(\textbf{E}_{text}(G_{\varphi{_1}}^{v_i}),\textbf{E}_{text}(obj))$ is the cosine similarity between the embedding of object $obj$ and the embedding at node $v_i$ of the template $G_{\varphi{_1}}$; \textbf{c} is a tuneable parameter that balances selecting high-cosine similarity objects vs relevant but unseen objects; \textbf{$N_t$} is the total number of object selection trails; \textbf{$N_{t} ^{obj}$} is the number of trails where object $obj$ was selected.

\subsection{Off-policy Updates via Counterfactual Experience} \label{sec:learning}
Concerning off-policy updates, when an episode is run with an epsilon greedy behavior policy on the given LTL formula, the Q-value function of each auxiliary task will also be updated as if their corresponding formula was the current objective. Assuming an action \emph{a} is taken in state \emph{s} resulting in a new state $s^{'}$, to update a specific $Q_{\varphi}$, the reward that would have been observed during that transition if the agent's objective was $\varphi$ is computed. To achieve this, $\varphi$ is progressed through the new state $s^{'}$; if the resulting formula $\varphi^{'}$ is \emph{true} the reward is 1, and 0 otherwise. $Q_\varphi$ is then updated using the following  rule:

\begin{equation}
    Q_{\varphi}(s,a) \leftarrow Q_{\varphi}(s,a) + \alpha(r + \gamma \max_{a^{'}} Q_{\varphi^{'}} (s^{'},a^{'}) - Q_{\varphi}(s,a))
\end{equation}

As shown in LPOPL \cite{rodrigoLPOPL} learning is globally optimal as $Q_{\varphi}(s,a)$ is updated with the maximum of every action $a^{'}$ from its progressed subtask $Q_{\varphi^{'}} (s^{'},a^{'})$.

\section{\texorpdfstring{\protect\hypertarget{experiment_section}{Experiments}}{}}
The task specification used in our experiments was a food preparation task where the agent had to go to the \textbf{kitchen cabinet}, obtain a \textbf{cooking pot}, obtain \textbf{seasoning}, then go to the \textbf{fridge}, obtain \textbf{chicken}, and finally go to the \textbf{stove}. In HomeGrid, this task corresponds to visiting the right cells in the correct order. The LTL formula below represents this task using the atomic propositions that represent each of the relevant objects:
\begin{equation}
   \lozenge (C \land \lozenge (P \land \lozenge (I \land \lozenge (F \land \lozenge (H \land \lozenge Y)))))
\end{equation}

See Figure \ref{fig:environment} for a description of our environment HomeGrid. We find that a good heuristic for choosing the minimum number of object clusters is the number of distinct exploration zones in the environment. For HomeGrid, this is four (4) since there are four distinct useful exploration regions namely "Kitchen", "Bathroom", "Living room" and "Bedroom". We evaluate our approach with the following three conditions:

\begin{enumerate}   
    \item \textbf{Ours}: Given our food preparation task $\varphi_1$  we generate \textbf{20 auxiliary tasks} following our approach. We learn these tasks simultaneously with $\varphi_1$ with a behavior policy epsilon greedy on $\varphi_1$, directing exploration towards more relevant experiences for $\varphi_1$.

    \item \textbf{Baseline 1}: To demonstrate that tasks generated by TaskExplore uniquely leverage the directed experience of a single-task curriculum, we repeat the approach described above, replacing the behavior policy with a random one that explores more widely.
    
    \item \textbf{Baseline 2}: To show that tasks generated by TaskExplore more relevantly benefit from directed exploration experience than the general set of possible tasks, we generate \textbf{20 auxiliary tasks} by randomly sampling sequential tasks from the set of propositions in our environment, as typically done in prior works \cite{vaezipoor2021ltl2action,liu2022skill}. These tasks are sampled to have the same length of propositions as the given task which places them in the same level of difficulty. We learn these tasks simultaneously with $\varphi_1$, using a behavior policy epsilon greedy on $\varphi_1$.
  
\end{enumerate}

\begin{figure*}
\centering
 \makebox[\textwidth][c]{\includegraphics[width=1\textwidth]{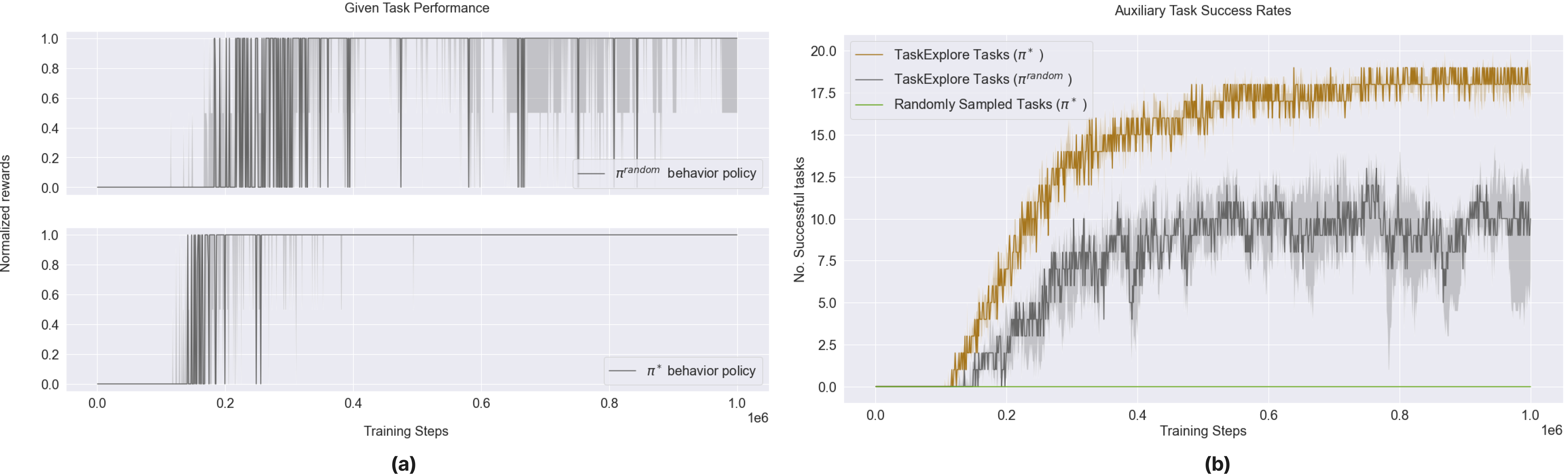}}%
    % \vspace*{-5mm}
  \caption{Figure (a) shows the normalized discounted reward obtained by the agent on the given task as it learns to solve it simultaneously with TaskExplore generated auxiliary tasks using a random behavior policy ($\pi^{random}$) and an epsilon greedy behavior policy ($\pi^*$). Figure (b) shows the task success rate on auxiliary tasks as learning progresses. Learning TaskExplore generated tasks while using epsilon greedy ($\pi^*$) behavior policy on the given task significantly outperforms all other baselines. All results are normalized over 7 different seeded runs}
  \label{fig:results}
     \vspace*{-2mm}
\end{figure*}
\section{Results and Discussion}

Figure \ref{fig:results} presents the results from our experiments, which highlight several interesting properties of the auxiliary task set developed by TaskExplore. Firstly, in Figure \ref{fig:results}a, we see that our approach which simultaneously learns to solve auxiliary tasks with a behavior policy $\epsilon$-greedy on the given task does not adversely affect performance on the given task. However, performance on the given task deteriorates when using a random behavior policy, highlighting the relevance of directed exploration. 

Intuitively, learning with a random behavior policy gathers more diverse experiences which should benefit multiple auxiliary tasks more than the  experience gathered during directed exploration. However, our results in \ref{fig:results}b  show that the auxiliary tasks developed by TaskExplore maximally leverage that constrained experience and actually performs better than when using a random behavior policy. This is because the generated tasks are contextually similar to the given task and share similar underlying exploration requirements. 

\begin{figure*}
\centering
 \makebox[\textwidth][c]{\includegraphics[width=0.8\textwidth]{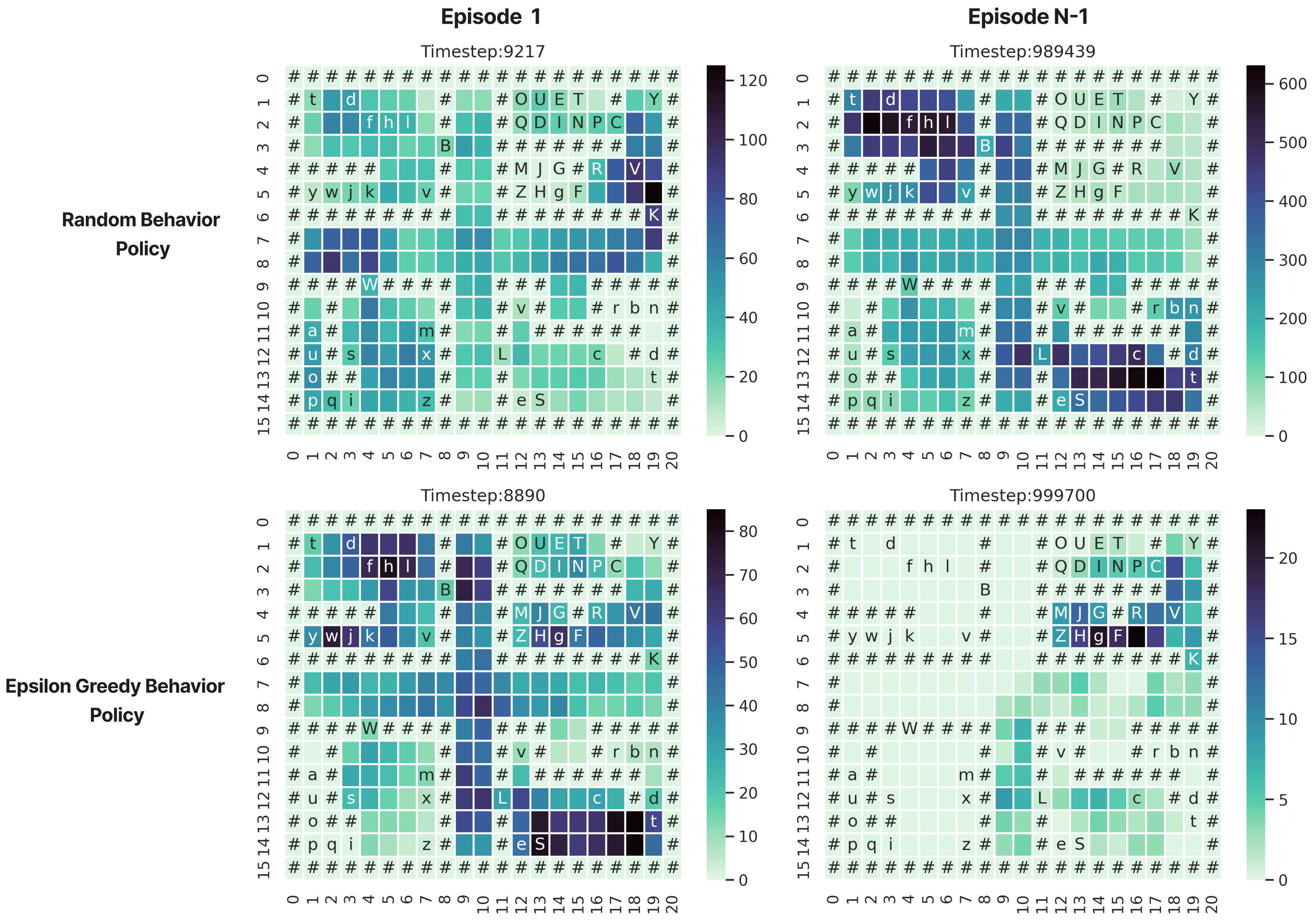}}%
  \caption{This figure shows a sample exploration heatmap of the random and e-greedy behavior policies while learning the given food preparation task with counterfactual updates on TaskExplore generated auxiliary tasks. The agent starts each episode in cell (x-axis=14,y-axis=13) and darker cell colors correlates to the number of times the agent visited that cell. This is an ablative diagram that helps visualize the beneficial effects of directed exploration in complex environments and how this directed experience can benefit contextually similar auxiliary tasks that share the same underlying exploration requirements, more than an exploration strategy that may produce more diverse experiences.}
  \label{fig:exploration}
  \vspace*{-2mm}
\end{figure*}

Figure \ref{fig:exploration} presents further insights into this phenomenon. It shows a sample exploration heatmap of a single complete run of our experiment learning the food preparation task with random and epsilon-greedy behaviour policies. In the first episode both behaviour policies explore widely, however as learning progresses the epsilon-greedy policy leads to more directed and constrained experiences that focuses on the Kitchen exploration zone. The random behaviour policy on the other hand still explores widely, not paying much attention to the kitchen exploration zone it needs to be focusing on to make progress on relevant auxiliary tasks.

% This is a useful property as it shows that directed exploration towards solving a specific task–which is often vital in environments with complex state spaces–is counter-intuitively better for our auxiliary tasks than widely exploring.

Finally, the results  in \ref{fig:results}b show that developing a curriculum of randomly sampled tasks from the general distribution space of possible tasks in the environment cannot leverage directed exploration experience as well as auxiliary tasks developed by TaskExplore. 

\section{Conclusion} 
This paper introduced an approach to how agents might automatically generate expressive temporally extended auxiliary tasks that can maximally leverage the directed experience of a single-task curriculum in object-centric environments. This approach to auxiliary task generation is particularly valuable in the lifelong learning setting, as agents can generate and solve new tasks from constrained datasets. In the spirit of reusing computation, a policy bank of these policies can be saved and reused to accelerate learning future tasks. Modern vision-language models (VLMs) that detect open vocabulary objects in real world environemnts can be employed in future work to relax TaskExplore's dependence on a predefind set of object propositions from which tasks can be expressed and labelling functions that map states to proposition truth values, 

\bibliographystyle{ieeetr}
\bibliography{references}

\begin{thebibliography}{10}

\bibitem{mnih2015human}
V.~Mnih, K.~Kavukcuoglu, D.~Silver, A.~A. Rusu, J.~Veness, M.~G. Bellemare, A.~Graves, M.~Riedmiller, A.~K. Fidjeland, G.~Ostrovski, {\em et~al.}, ``Human-level control through deep reinforcement learning,'' {\em nature}, vol.~518, no.~7540, pp.~529--533, 2015.

\bibitem{schulman2017proximal}
J.~Schulman, F.~Wolski, P.~Dhariwal, A.~Radford, and O.~Klimov, ``Proximal policy optimization algorithms,'' {\em arXiv preprint arXiv:1707.06347}, 2017.

\bibitem{berner2019dota}
C.~Berner, G.~Brockman, B.~Chan, V.~Cheung, P.~Debiak, C.~Dennison, D.~Farhi, Q.~Fischer, S.~Hashme, C.~Hesse, {\em et~al.}, ``Dota 2 with large scale deep reinforcement learning,'' {\em arXiv preprint arXiv:1912.06680}, 2019.

\bibitem{lin1992self}
L.-J. Lin, ``Self-improving reactive agents based on reinforcement learning, planning and teaching,'' {\em Machine learning}, vol.~8, pp.~293--321, 1992.

\bibitem{schaul2015prioritized}
T.~Schaul, J.~Quan, I.~Antonoglou, and D.~Silver, ``Prioritized experience replay,'' {\em arXiv preprint arXiv:1511.05952}, 2015.

\bibitem{andrychowicz2017hindsight}
M.~Andrychowicz, F.~Wolski, A.~Ray, J.~Schneider, R.~Fong, P.~Welinder, B.~McGrew, J.~Tobin, O.~Pieter~Abbeel, and W.~Zaremba, ``Hindsight experience replay,'' {\em Advances in neural information processing systems}, vol.~30, 2017.

\bibitem{icarte2022reward}
R.~T. Icarte, T.~Q. Klassen, R.~Valenzano, and S.~A. McIlraith, ``Reward machines: Exploiting reward function structure in reinforcement learning,'' {\em Journal of Artificial Intelligence Research}, vol.~73, pp.~173--208, 2022.

\bibitem{icarte2018using}
R.~T. Icarte, T.~Klassen, R.~Valenzano, and S.~McIlraith, ``Using reward machines for high-level task specification and decomposition in reinforcement learning,'' in {\em International Conference on Machine Learning}, pp.~2107--2116, PMLR, 2018.

\bibitem{rodrigoLPOPL}
R.~Toro~Icarte, T.~Q. Klassen, R.~Valenzano, and S.~A. McIlraith, ``Teaching multiple tasks to an rl agent using ltl,'' in {\em Proceedings of the 17th International Conference on Autonomous Agents and MultiAgent Systems}, AAMAS '18, p.~452–461, International Foundation for Autonomous Agents and Multiagent Systems, 2018.

\bibitem{pnueliLTL}
A.~Pnueli, ``The temporal logic of programs,'' in {\em Proceedings of the 18th Annual Symposium on Foundations of Computer Science}, SFCS '77, p.~46–57, IEEE Computer Society, 1977.

\bibitem{littman2017environment}
M.~L. Littman, U.~Topcu, J.~Fu, C.~Isbell, M.~Wen, and J.~MacGlashan, ``Environment-independent task specifications via gltl,'' {\em arXiv preprint arXiv:1704.04341}, 2017.

\bibitem{bacchus2000using}
F.~Bacchus and F.~Kabanza, ``Using temporal logics to express search control knowledge for planning,'' {\em Artificial intelligence}, vol.~116, no.~1-2, pp.~123--191, 2000.

\bibitem{watkins1992q}
C.~J. Watkins and P.~Dayan, ``Q-learning,'' {\em Machine learning}, vol.~8, pp.~279--292, 1992.

\bibitem{narvekar2020curriculum}
S.~Narvekar, B.~Peng, M.~Leonetti, J.~Sinapov, M.~E. Taylor, and P.~Stone, ``Curriculum learning for reinforcement learning domains: A framework and survey,'' {\em The Journal of Machine Learning Research}, vol.~21, no.~1, pp.~7382--7431, 2020.

\bibitem{narvekar2016source}
S.~Narvekar, J.~Sinapov, M.~Leonetti, and P.~Stone, ``Source task creation for curriculum learning,'' in {\em Proceedings of the 2016 international conference on autonomous agents \& multiagent systems}, pp.~566--574, 2016.

\bibitem{silva2018object}
F.~L.~D. Silva and A.~H.~R. Costa, ``Object-oriented curriculum generation for reinforcement learning,'' in {\em Proceedings of the 17th international conference on autonomous agents and multiagent systems}, pp.~1026--1034, 2018.

\bibitem{thrun1998lifelong}
S.~Thrun, ``Lifelong learning algorithms.,'' {\em Learning to learn}, vol.~8, pp.~181--209, 1998.

\bibitem{diuk2008object}
C.~Diuk, A.~Cohen, and M.~L. Littman, ``An object-oriented representation for efficient reinforcement learning,'' in {\em Proceedings of the 25th international conference on Machine learning}, pp.~240--247, 2008.

\bibitem{brafman2018ltlf}
R.~Brafman, G.~De~Giacomo, and F.~Patrizi, ``Ltlf/ldlf non-markovian rewards,'' in {\em Proceedings of the AAAI conference on artificial intelligence}, vol.~32, 2018.

\bibitem{camacho2017non}
A.~Camacho, O.~Chen, S.~Sanner, and S.~A. McIlraith, ``Non-markovian rewards expressed in ltl: guiding search via reward shaping,'' in {\em Tenth annual symposium on combinatorial search}, 2017.

\bibitem{thiebaux2006decision}
S.~Thi{\'e}baux, C.~Gretton, J.~Slaney, D.~Price, and F.~Kabanza, ``Decision-theoretic planning with non-markovian rewards,'' {\em Journal of Artificial Intelligence Research}, vol.~25, pp.~17--74, 2006.

\bibitem{liu2022skill}
J.~X. Liu, A.~Shah, E.~Rosen, G.~Konidaris, and S.~Tellex, ``Skill transfer for temporally-extended task specifications,'' {\em arXiv preprint arXiv:2206.05096}, 2022.

\bibitem{radford2018improving}
A.~Radford, K.~Narasimhan, T.~Salimans, I.~Sutskever, {\em et~al.}, ``Improving language understanding by generative pre-training,'' 2018.

\bibitem{raffel2020exploring}
C.~Raffel, N.~Shazeer, A.~Roberts, K.~Lee, S.~Narang, M.~Matena, Y.~Zhou, W.~Li, and P.~J. Liu, ``Exploring the limits of transfer learning with a unified text-to-text transformer,'' {\em The Journal of Machine Learning Research}, vol.~21, no.~1, pp.~5485--5551, 2020.

\bibitem{devlin2018bert}
J.~Devlin, M.-W. Chang, K.~Lee, and K.~Toutanova, ``Bert: Pre-training of deep bidirectional transformers for language understanding,'' {\em arXiv preprint arXiv:1810.04805}, 2018.

\bibitem{ouyang2022training}
L.~Ouyang, J.~Wu, X.~Jiang, D.~Almeida, C.~L. Wainwright, P.~Mishkin, C.~Zhang, S.~Agarwal, K.~Slama, A.~Ray, {\em et~al.}, ``Training language models to follow instructions with human feedback,'' {\em arXiv preprint arXiv:2203.02155}, 2022.

\bibitem{ni2021sentence}
J.~Ni, G.~H. {\'A}brego, N.~Constant, J.~Ma, K.~B. Hall, D.~Cer, and Y.~Yang, ``Sentence-t5: Scalable sentence encoders from pre-trained text-to-text models,'' {\em arXiv preprint arXiv:2108.08877}, 2021.

\bibitem{lloyd1982least}
S.~Lloyd, ``Least squares quantization in pcm,'' {\em IEEE transactions on information theory}, vol.~28, no.~2, pp.~129--137, 1982.

\bibitem{macqueen1965some}
J.~MacQueen, ``Some methods for classification and analysis of multivariate observations,'' in {\em Proc. 5th Berkeley Symposium on Math., Stat., and Prob}, p.~281, 1965.

\bibitem{arthur2006k}
D.~Arthur and S.~Vassilvitskii, ``k-means++: The advantages of careful seeding,'' tech. rep., Stanford, 2006.

\bibitem{vaezipoor2021ltl2action}
P.~Vaezipoor, A.~C. Li, R.~A.~T. Icarte, and S.~A. Mcilraith, ``Ltl2action: Generalizing ltl instructions for multi-task rl,'' in {\em International Conference on Machine Learning}, pp.~10497--10508, PMLR, 2021.

\bibitem{auer2002using}
P.~Auer, ``Using confidence bounds for exploitation-exploration trade-offs,'' {\em Journal of Machine Learning Research}, vol.~3, no.~Nov, pp.~397--422, 2002.

\end{thebibliography}

\end{document}